\documentclass[twoside,11pt]{article}

\usepackage{jmlr2e}

\usepackage{url}

\hypersetup{
  colorlinks=true,
	citecolor=blue!50!blue,
  linkcolor=blue!50!blue,
  urlcolor=blue!70!blue
}

\usepackage{times}
\usepackage{graphicx} % more modern
\usepackage{rotating}
\usepackage{multirow}
\usepackage{natbib}
\usepackage{amssymb}
\usepackage{amsmath}
\usepackage{mathtools}
\usepackage{algorithm}
\usepackage{algorithmic}
\usepackage{multirow}
\usepackage{tcolorbox}
\usepackage{tabularx}
\usepackage{array}
\usepackage{colortbl}
\usepackage{fancybox}
\usepackage{tikz}

\tcbuselibrary{skins}

\usepackage[utf8]{inputenc} % allow utf-8 input
\usepackage[T1]{fontenc}    % use 8-bit T1 fonts
\usepackage{hyperref}       % hyperlinks
\usepackage{url}            % simple URL typesetting
\usepackage{booktabs}       % professional-quality tables
\usepackage{amsfonts}       % blackboard math symbols
\usepackage{nicefrac}       % compact symbols for 1/2, etc.
\usepackage{microtype}      % microtypography
\usepackage{amssymb}
\usepackage{amsmath}
\usepackage{mathtools}
\usepackage{bigdelim} 
\usepackage{algorithm}
\usepackage{algorithmic}
\usepackage{color}
\usepackage{verbatim}
\usepackage{siunitx}

\tcbuselibrary{skins}
\usepackage{fancybox}
\usepackage{tikz}
\usepackage{multirow}
\usepackage{tcolorbox}
\usepackage{tabularx}
\usepackage{array}
\usepackage{colortbl}
\usepackage{subcaption}

\usepackage{hyperref}

\tcbset{tab2/.style={fonttitle=\bfseries\footnotesize,fontupper=\footnotesize,
colback=blue!2!white,colframe=blue!75!black,colbacktitle=red!40!white,
coltitle=black,center title,freelance}}

\usepackage{tikz}
\usetikzlibrary{positioning}
\newcommand\annotate[3][below]{%
\tikz[baseline=0pt,trim left=(@.west), trim right=(@.east)]{%
  \node [draw=orange!50, top color=white, bottom color=orange!25, rounded corners=2pt, inner xsep=1pt,anchor=base](@){#2};
  \node [font=\footnotesize, align=center,inner ysep=3pt, #1=1cm/4 of @] (@') {#3};
  \draw[cyan,-stealth](@)--(@');}%
  \ignorespaces%
}

\usepackage[font=small,skip=5.5pt]{caption}

\ShortHeadings{Forecasting Individualized Disease Trajectories}{Alaa and van der Schaar}
\firstpageno{1}

\begin{document}

\title{Forecasting Individualized Disease Trajectories using\\ Interpretable Deep Learning}

\author{\name Ahmed M. Alaa \email ahmedmalaa@ucla.edu \\
       \addr Electrical and Computer Engineering Department\\
       University of California, Los Angeles\\
       Los Angeles, CA 90095-1594, USA
       \AND
       \name Mihaela van der Schaar \email mihaela@ee.ucla.edu \\
       \addr Electrical Engineering Department\\
       University of California, Los Angeles\\
       Los Angeles, CA 90095-1594, USA}

\editor{XXXXX}

\maketitle

\begin{abstract}%   <- trailing '%' for backward compatibility of .sty file
Disease progression models are instrumental in {\it predicting} individual-level health trajectories and {\it understanding} disease dynamics. Existing models are capable of providing either accurate predictions of patients' prognoses or clinically interpretable representations of disease pathophysiology, but not both. In this paper, we develop the {\it phased attentive state space} (PASS) model of disease progression, a deep probabilistic model that captures complex representations for disease progression while maintaining clinical interpretability. Unlike Markovian state space models which assume memoryless dynamics, PASS uses an attention mechanism to induce "memoryful" state transitions, whereby repeatedly updated attention weights are used to focus on past state realizations that best predict future states. This gives rise to complex, non-stationary state dynamics that remain interpretable through the generated attention weights, which designate the relationships between the realized state variables for {\it individual} patients. PASS uses {\it phased} LSTM units (with time gates controlled by parametrized oscillations) to generate the attention weights in continuous time, which enables handling irregularly-sampled and potentially missing medical observations. Experiments on data from a real-world cohort of patients show that PASS successfully balances the tradeoff between accuracy and interpretability: it demonstrates superior predictive accuracy and learns insightful individual-level representations of disease progression.
\end{abstract}


\section{Introduction}
\label{SSec1}
Chronic diseases -- such as cardiovascular disease, cancer and diabetes -- progress slowly throughout a patient's lifetime, causing increasing burden to the patients, their carers, and the healthcare delivery system \cite{sevick2007patients}. Modern electronic health records (EHR) keep track of individual patients' disease progression trajectories through follow-up data sequences of the form \mbox{\footnotesize $(\boldsymbol{X}_1, . . . , \boldsymbol{X}_t)$}, where \mbox{\footnotesize $\boldsymbol{X}_t$} is a set of clinical observations collected for the patient at time \mbox{\footnotesize $t$}. The advent of EHRs\footnote{\url{https://www.healthit.gov/sites/default/files/briefs/}} provides an opportunity for building models of disease progression that can fulfill two central goals of healthcare delivery systems:
\begin{itemize}
\item \textbf{\underline{Goal A}: \textit{Predicting} individual-level disease trajectories.} 
\item \textbf{\underline{Goal B}: \textit{Understanding} disease progression mechanisms.}   
\end{itemize} 
{\bf Goal A} entails the {\it supervised} problem of predicting future clinical observations \mbox{\footnotesize $(\boldsymbol{X}_{t+1},\boldsymbol{X}_{t+2},.\,.\,.)$} on the basis of past observations \mbox{\footnotesize $(\boldsymbol{X}_1,.\,.\,., \boldsymbol{X}_t)$}. {\bf Goal B} entails the {\it unsupervised} problem of discovering clinically-interpretable latent structures that explain the mechanisms underlying disease progression. Both goals {\bf A} and {\bf B} are entangled. This is because accurate predictions need to be transparent and interpretable in order to ensure their actionability, whereas interpretable representations explaining disease progression can only be trustworthy if they possess high predictive power. 

\begin{figure*}[t!] 
    \centering
    \begin{subfigure}[t]{0.32\textwidth}
        \centering
        \includegraphics[width=1.75in]{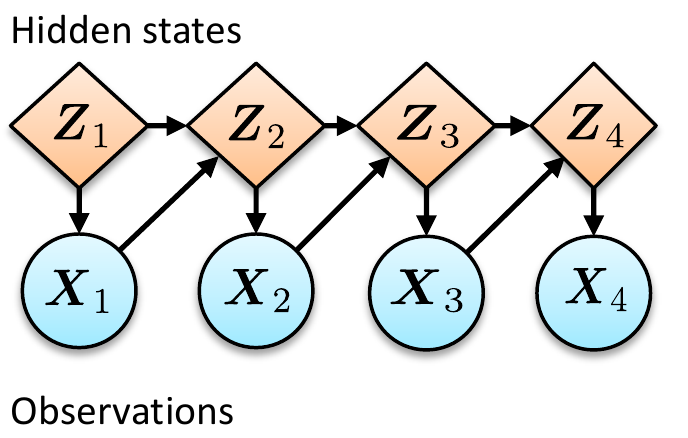}
        \caption{\footnotesize RNN}
				\label{fig1a}
    \end{subfigure}%
    ~ 
    \begin{subfigure}[t]{0.32\textwidth}
        \centering
        \includegraphics[width=1.75in]{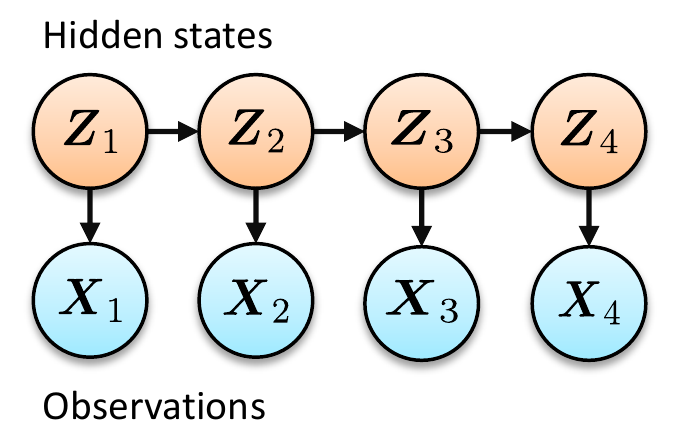}
        \caption{\footnotesize HMM}
				\label{fig1b}
    \end{subfigure}
		 ~ 
    \begin{subfigure}[t]{0.32\textwidth}
        \centering
        \includegraphics[width=1.75in]{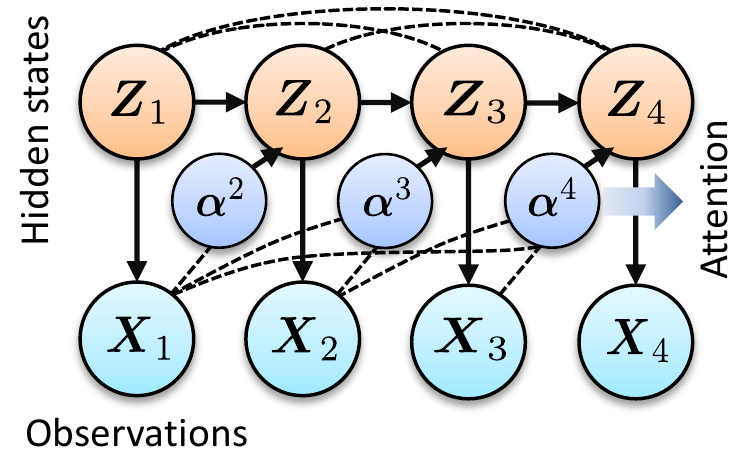}
        \caption{\footnotesize Attentive state space}
				\label{fig1c}
		 \end{subfigure}		
				 ~ 
    \begin{subfigure}[t]{0.64\textwidth}
        \centering
        \includegraphics[width=3.25in]{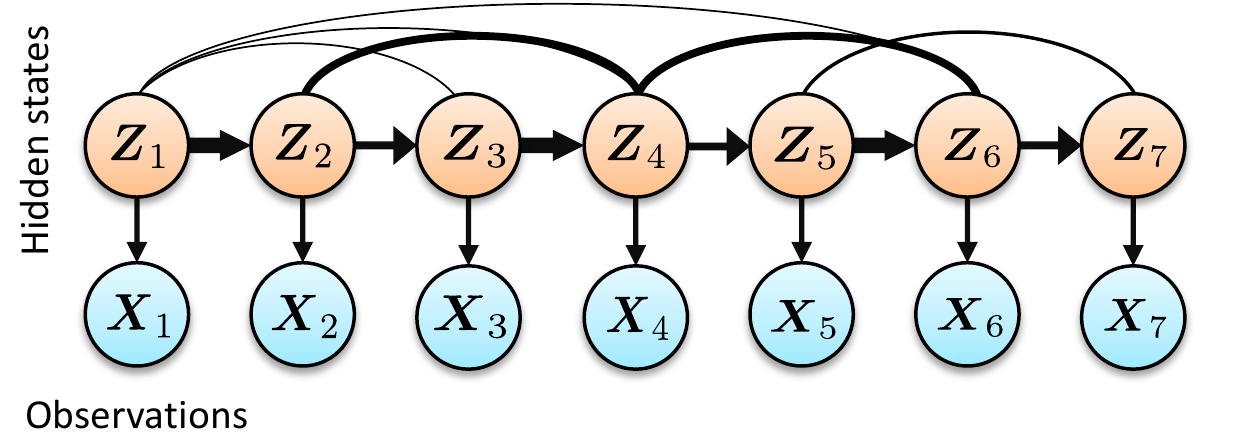}
        \caption{\footnotesize Unrolled graphical depiction for an attentive state space}
				\label{fig1c2}
		 \end{subfigure}		
    \caption{\footnotesize Depictions for different models of sequential data: (a) Graphical model for an RNN. $\Diamond$ denotes a deterministic intermediate representation, (b) Graphical model for an HMM. $\bigcirc$ denotes probabilistic states, (c) Graphical model for the proposed attentive state space model. (c) Unrolled graphical depiction for an attentive state space. Thickness of the arrows reflect the attention weights.}
\end{figure*}

Unfortunately, as a consequence of the inherent tension between model {\it accuracy} and {\it interpretability} \cite{lipton2016mythos}, most existing models of disease progression fulfill either {\bf Goal A} or {\bf Goal b}, but not both. State-of-the-art prediction performance is achieved by models based on {\it recurrent neural networks} (RNN) \cite{lim2018disease,lipton2015learning,choi2016doctor}. RNN-based models are often used for sequence prediction (or sequence labeling), where they are trained to estimate the predictive distribution \mbox{\footnotesize $P(\boldsymbol{X}_t \,|\,\boldsymbol{X}_{t-1}, . . . ,\boldsymbol{X}_{1})$} by propagating a sequence of hidden states \mbox{\footnotesize $(\boldsymbol{Z}_1, . . . , \boldsymbol{Z}_t)$} through intermediate deterministic mappings (Figure \ref{fig1a}). Unfortunately, an RNN is of a "black-box" nature since its hidden states \mbox{\footnotesize $(\boldsymbol{Z}_1, . . . , \boldsymbol{Z}_t)$} do not explicitly map to clinically meaningful states of disease progression. On the contrary, {\it state space} approaches based on Hidden Markov Models (HMM) provide a natural interpretation of a disease trajectory as a sequence of transitions between latent "progression stages" \mbox{\footnotesize $(\boldsymbol{Z}_1, . . . , \boldsymbol{Z}_t)$} (Figure \ref{fig1b}), each of which corresponds to a clinically distinguishable disease state \cite{alaa2016hidden,liu2015efficient,wang2014unsupervised}. However, the interpretability of HMMs comes at a price. That is, while RNNs can in principle approximate {\it any} dynamical system, an HMM is limited to memoryless Markovian state dynamics, which greatly undermine its predictive performance.

{\bf Our Contribution}\,\,\, In this paper, we develop a deep probabilistic model of disease progression that capitalizes on both the predictive power of RNNs and the interpretable nature of state space models to fulfill {\bf Goals A} and {\bf B}. Our model maintains the probabilistic structure of a state space representation, which decouples {\it emission} and {\it transition} distributions, but uses an RNN to model a flexible non-Markovian state transition dynamic \mbox{\footnotesize $P(\boldsymbol{Z}_t\,|\,\boldsymbol{Z}_{t-1}, . . . , \boldsymbol{Z}_1)$} that allows future states to depend on all past states. To model state transitions, we use an attention mechanism whereby the RNN generates a (repeatedly updated) set of attention weights \mbox{\footnotesize $(\boldsymbol{\alpha}^1, . . . , \boldsymbol{\alpha}^t)$} that designate the (relative) influence that past state realizations \mbox{\footnotesize $(\boldsymbol{Z}_1, . . . , \boldsymbol{Z}_t)$} have on the transition probabilities to the future state \mbox{\footnotesize $\boldsymbol{Z}_{t+1}$}. Our model for state transitions can be summarized as follows:
\begin{align}
(\boldsymbol{\alpha}_1^t, . . . , \boldsymbol{\alpha}_{t}^t) &= {\bf RNN}(\boldsymbol{X}_1, . . . , \boldsymbol{X}_t), \nonumber \\
P(\boldsymbol{Z}_{t+1} = z) &= \sum_{i \leq t} \boldsymbol{\alpha}_i^t \, \times \, P(\boldsymbol{Z}_t = z\,|\,\boldsymbol{Z}_i = z^{\prime}). \nonumber
\end{align}
The model described above, which we call an {\it attentive state space} model, uses a set of RNN-generated attention weights to induce a time-varying Markov blanket for the state variable \mbox{\footnotesize $\boldsymbol{Z}_{t}$}. This Markov blanket changes for every new state transition, putting more or less attention on previous state realizations depending on the patient's clinical history. If we restrict the attention weights to be binary ($\boldsymbol{\alpha}_i^t \in \{0,1\}, \forall i \leq t$), the attentive state space becomes a {\it variable-order} Markov model that decides the extent of memory involved in state transitions in every time step depending on the patient's current context. This allows for realistic non-stationary and time-inhomogeneous dynamics that are implicitly captured by an RNN but could not be possibly modeled with an HMM. The attentive state representation is clinically interpretable because the complex state dynamics that it captures are fully explicable through the attention weights, which indicates the extent to which past clinical events contribute to future state realizations. Figure \ref{fig1c} provides a formal graphical model for our attentive state space representation. Figure \ref{fig1c2} shows an unrolled graphical depiction of the model for a particular exemplary patient, highlighting the time-varying nature of the attention weights and their straightforward interpretational benefit.   

Because EHR data comprises irregularly-sampled and asynchronous observations gathered only at the times when the patient visits a hospital, we use the {\it phased} LSTM units introduced in \cite{neil2016phased}, with time gates controlled by parametrized oscillations, in order to generate the attention weights at arbitrary time instances. Thus, we call our model a {\it phased attentive state space} (PASS) model. A detailed description of the construction of the PASS model is provided in Section \ref{Sec3}. A detailed comparison between PASS and related models can be found in Section \ref{Sec2}. 

Indeed, state inference and parameter estimation of PASS is nontrivial since non-Markovianity hinders the application of conventional backward message passing algorithms \cite{alaa2016hidden,dai2016recurrent}. In Section \ref{Sec4}, we show that PASS can be re-parameterized as a non-stationary dynamic Bayesian network, for which conventional forward message passing algorithms can be implemented with a complexity resembling that of the forward filtering algorithm used for HMMs. We conduct parameter estimation via a variant of the Expectation-Maximization algorithm. In Section \ref{Sec5}, we conduct experiments on data from a real-world longitudinal cohort of more than 10,000 Cystic Fibrosis (CF) patients. Our experiments show that PASS successfully balances the tradeoff between accuracy and interpretability: it demonstrates superior predictive accuracy and learns insightful individual-level representations of disease progression. In particular, we show that PASS learns meaningful population-level CF progression stage, and that the attention weights can inform treatment decisions on the level of individual patients.    

\section{A Phased Attentive State-space Model of Disease Progression}
\label{Sec3}
We model the progression of a target chronic disease using longitudinal EHR data for patients who have developed, or are at risk of developing such disease. We start by describing the model variables in Section \ref{Sec31}, and then we develop the attentive state dynamics in Sections \ref{Sec32} and \ref{Sec33}. 
     
\subsection{Model Variables and Notation}
\label{Sec31} 
{\bf Structure of the EHR data}\,\, A patient's EHR record, denoted as \mbox{\footnotesize $\boldsymbol{\mathcal{D}}$}, is a collection of timestamped follow-up data gathered during repeated, irregularly-spaced hospital visits, in addition to static features (e.g., genetic variables). We represent a given patient's EHR record as follows:  
\begin{equation}
\boldsymbol{\mathcal{D}} = {\annotate{$\{\boldsymbol{Y}\}$}{$\mbox{\footnotesize \bf Static features}$}} \cup \{(\boldsymbol{X}_{m},{\annotate{$t_{m}$}{$\mbox{\footnotesize \bf Visit times}$}})\}^{M}_{m=1}, 
\label{eq00}
\end{equation}    
where \mbox{\footnotesize $\boldsymbol{Y}$} is the static features' vector, \mbox{\footnotesize $\boldsymbol{X}_{m}$} is the follow-up data collected in the \mbox{\footnotesize $m^{th}$} hospital visit, \mbox{\footnotesize $t_m$} is the time of the \mbox{\footnotesize $m^{th}$} visit, and \mbox{\footnotesize $M$} is the total number of hospital visits. (The time-horizon \mbox{\footnotesize $t$} is taken to be the patient's chronological age.) The follow-up data \mbox{\footnotesize $\boldsymbol{X}_{m}$} comprises information on biomarkers and clinical events, such as treatments and diagnoses of comorbidities. (Refer to the Appendix for a more elaborate discussion on the type of follow-up data collected in EHRs.) An EHR dataset \mbox{\footnotesize $\{\boldsymbol{\mathcal{D}}^{(i)}\}_{i=1}^N$} is an assembly of records for \mbox{\footnotesize $N$} independent patients. 

{\bf Disease progression stages}\,\, We assume that the target disease evolves through \mbox{\footnotesize $D$} different {\it progression stages}. Each stage corresponds to a distinct level of disease severity that manifests through the follow-up data. We model the evolution of progression stages via a (continuous-time) stochastic process \mbox{\footnotesize $\boldsymbol{Z}(t)$} of the following form: 
\begin{align}
\boldsymbol{Z}(t) = \sum_{n \in \mathbb{N}_+}\,\tilde{Z}_n\,\cdot\,{\bf 1}_{\{T_{n}\, < \, t \, \leq \, T_{n+1}\}},\,\, \tilde{Z}_n \in \{1,\,.\,.\,.,D\}, \nonumber   
%\label{eq0}
\end{align}
where \mbox{\footnotesize $\{T_n\}_n$} is the sequence of {\it onsets} for the realized progression stages, and \mbox{\footnotesize $\tilde{Z}_n$} is the progression stage occupying the interval \mbox{\footnotesize $(T_{n},T_{n+1}]$}. We assume that \mbox{\footnotesize $\boldsymbol{Z}(0) = 1$} (i.e., \mbox{\footnotesize $\tilde{Z}_1 = 1$}) almost surely, with stage 1 being the {\it asymptomatic} stage designating "healthy" patients. The sequence \mbox{\footnotesize $\{\boldsymbol{Z}_m\}_m$} is the {\it embedded} discrete-time process induced by \mbox{\footnotesize $\boldsymbol{Z}(t)$} at the hospital visit times \mbox{\footnotesize $\{t_m\}_m$}, i.e. \mbox{\footnotesize $\boldsymbol{Z}_m = \boldsymbol{Z}(t_m)$}. %Thus, the observable EHR record \mbox{\footnotesize $\boldsymbol{\mathcal{D}}$} is modulated by the sequence \mbox{\footnotesize $\{\boldsymbol{Z}_m\}_m$}.     

\subsection{Attentive State Space Representation}
\label{Sec32}
We adopt a state space representation for the disease progression process, with the state space being the set of all stages of progression \mbox{\footnotesize $\{1,.\,.\,.\,,D\}$}. The states' sequence \mbox{\footnotesize $\{\boldsymbol{Z}_m\}_m$} is {\it hidden} whereas the EHR data \mbox{\footnotesize $\boldsymbol{\mathcal{D}}$} is {\it observed}. We consider a graphical model that defines probabilistic dependencies between \mbox{\footnotesize $\{\boldsymbol{Z}_m\}_m$} and \mbox{\footnotesize $\boldsymbol{\mathcal{D}}$} through the following factorization of {\it emission} and {\it transition} distributions: 
\begin{align}
&P(\{\boldsymbol{Z}_m\}_m,\{\boldsymbol{X}_m\}_m\,|\,\boldsymbol{Y},\{t_m\}_m) = \nonumber \\ 
&\prod^{m}_{m^{\prime}=1} \underbrace{P(\boldsymbol{X}_{m^{\prime}}\,|\,\boldsymbol{Z}_{m^{\prime}})}_{\mbox{\footnotesize \textbf{Emission}}}\,\cdot\,\underbrace{P(\boldsymbol{Z}_{m^{\prime}}\,|\,\mathcal{F}_{t_{m^{\prime}-1}})}_{\mbox{\footnotesize \textbf{Transition}}}, 
\label{eq01}
\end{align}
where \mbox{\footnotesize $\mathcal{F}_{t_{m^{\prime}}} = \left\{\boldsymbol{Y},(\boldsymbol{Z}_1,\boldsymbol{X}_1,t_{1}),.\,.\,.\,,(\boldsymbol{Z}_{m^{\prime}},\boldsymbol{X}_{m^{\prime}},t_{m^{\prime}})\right\}$} conveys all the information available in the model up to time \mbox{\footnotesize $t_{m^{\prime}}$}. We model the emission distribution in (\ref{eq01}) as a Gaussian distribution with state-specific parameters as follows: 
\begin{align}
P(\boldsymbol{X}_{m}\,|\,\boldsymbol{Z}_{m}=z) = \mathcal{N}(\boldsymbol{\mu}_z, \boldsymbol{\Sigma}_z),\, z \in \{1,.\,.\,.\,,D\}. 
\label{eq01Z}
\end{align} 
Binary variables are modeled with a Bernoulli state-specific distributions. The transition probability factor in (\ref{eq01}) assumes that the realized state at time \mbox{\footnotesize $t_{m}$} depends on the entire process history \mbox{\footnotesize $\mathcal{F}_{t_{m-1}}$}. To model \mbox{\footnotesize $P(\boldsymbol{Z}_{m^{\prime}}\,|\,\mathcal{F}_{t_{m^{\prime}-1}})$}, we first define a \mbox{\footnotesize $D \times D$} baseline Markov {\it generator matrix} \mbox{\footnotesize $\boldsymbol{\Lambda}$} as follows: 
\begin{align}
\boldsymbol{\Lambda} = \begin{bmatrix}
\, -\lambda_{12} & \lambda_{12} &  0  & \ldots & 0 \, \\
\, 0  & -\lambda_{23} & \lambda_{23} & \ldots & 0 \, \\
\, \vdots & \vdots & \ddots & \vdots & \vdots \, \\
\, 0 & \ldots & 0 & -\lambda_{_{D-1,D}} & \lambda_{_{D-1,D}} \, \\
\, 0 & \ldots &  0 & 0 & 0 \,
  \end{bmatrix}, 
\label{eq000}
\end{align}
\mbox{\footnotesize $\lambda_{ij} \geq 0,\,\, \forall i, j \in \{1,.\,.\,.\,,D\}$}, where \mbox{\footnotesize $\lambda_{ij}$} is a Markovian transition rate from state \mbox{\footnotesize $i$} to state \mbox{\footnotesize $j$}. \mbox{\footnotesize $\boldsymbol{\Lambda}$} is the transition rate matrix of a continuous-time Markov chain model on the state space \mbox{\footnotesize $\{1,.\,.\,.\,,D\}$}; its bidiagonal structure forces transitions to be permissible only between adjacent states (in an ascending order), with the last state (state \mbox{\footnotesize $D$}) being an absorbing state. This enforces the states in the set \mbox{\footnotesize $\{1,.\,.\,.\,,D\}$} to map properly to the disease progression stages (state \mbox{\footnotesize $1$} is the least severe stage and state \mbox{\footnotesize $D$} is the terminal stage of illness). We model the state transition probability \mbox{\footnotesize $P(\boldsymbol{Z}_m = z\,|\,\mathcal{F}_{t_{m-1}})$} by creating a "memoryful" version of the Markov chain model in (\ref{eq000}) through the following parametrization: 
\begin{align}
P(\boldsymbol{Z}_m = z\,|\,{\annotate{$\{(\boldsymbol{Z}_k = z_k,\boldsymbol{\alpha}^m_k,\Delta_k)\}_{k=1}^{m-1}$}{$\mbox{\footnotesize \bf Sufficient statistics}$}}) = \nonumber \\
\sum_{k=1}^{m-1}\,\,\, {\annotate{$\boldsymbol{\alpha}^m_k$}{$\mbox{\footnotesize \bf Attention weights}$}} \,\,\,(e^{\Delta_k \, \boldsymbol{\Lambda}})_{z,z_k}, %\nonumber
\label{eqy1}
\end{align}
where \mbox{\footnotesize $\Delta_k = t_{m}-t_{k}$} is the time interval between the \mbox{\footnotesize $k^{th}$} and the \mbox{\footnotesize $m^{th}$} hospital visits, \mbox{\footnotesize $\boldsymbol{\alpha}^m_k \in [0,1]$} is an {\it attention weight} assigned to the \mbox{\footnotesize $k^{th}$} visit, with \mbox{\footnotesize $\sum^{m-1}_{k=1}\boldsymbol{\alpha}^m_k = 1$}, and \mbox{\footnotesize $(e^{\Delta_k \, \boldsymbol{\Lambda}})_{z,z_k}$} is entry \mbox{\footnotesize $(z,z_k)$} of the exponentiation of matrix \mbox{\footnotesize $\boldsymbol{\Lambda}$}. The attention weights in (\ref{eqy1}) are generated via an attention function \mbox{\footnotesize $\boldsymbol{\varphi}$} with parameter \mbox{\footnotesize $\boldsymbol{\Theta}$} as follows:
\begin{align}
(\boldsymbol{\alpha}^m_1,.\,.\,.\,,\boldsymbol{\alpha}^m_{m-1}) = \boldsymbol{\varphi}(\boldsymbol{Y}, t_{m}, \{(\boldsymbol{X}_1,t_{1})\}_{j=1}^{m-1}; \boldsymbol{\Theta}).
\label{eq2}
\end{align}  
We call the representation in (\ref{eqy1}) an {\it attentive state space} representation. As shown in (\ref{eqy1}), the attentive representation starts with a baseline Markov chain, and creates memory in state transitions by weighting the baseline Markovian transition probabilities from all previous states, i.e. \mbox{\footnotesize $\{e^{\Delta_k \, \boldsymbol{Q}}\}^{m-1}_{k=1}$}, using a set of attention weights \mbox{\footnotesize $\{\boldsymbol{\alpha}^m_k\}^{m-1}_{k=1}$}. (The attention weights are generated on the basis of a patient's static features and follow-up data.) That is, instead of the memoryless Markovian dynamics in which a new state realization \mbox{\footnotesize $\boldsymbol{Z}_{m+1}$} is fully determined by the current realization \mbox{\footnotesize $\boldsymbol{Z}_{m}$}, the attentive state dynamics pay attention to all previous realizations in proportion to their attention weights. This gives rise to "memoryful", non-stationary, and time-inhomogeneous state transitions, whereby the time-varying Markov blanket (sufficient statistic) \mbox{\footnotesize $\{(\boldsymbol{Z}_k,\boldsymbol{\alpha}^m_k,\Delta_k)\}^{m-1}_{k=1}$} of every new state realization \mbox{\footnotesize $\boldsymbol{Z}_{m}$} determines which state realizations in the past matter most for the future.  

Similar to an HMM model, the factorization in (\ref{eq01}) decouples the transition and emission distributions by assuming that \mbox{\footnotesize $\boldsymbol{Z}_m$} $d$-separates \mbox{\footnotesize $\boldsymbol{X}_m$} from all other variables. This decoupling ensures that the clinical interpretability of the hidden states is maintained since each state is associated with a distinct emission distribution for the observed follow-up data. Moreover, regardless of the choice of the attention function \mbox{\footnotesize $\boldsymbol{\varphi}$}, the attentive state transition matrix \mbox{\footnotesize $\sum_{k=1}^{m-1}\,\,\, \boldsymbol{\alpha}^m_k \,e^{\Delta_k \, \boldsymbol{\Lambda}}$} will always be interpretable because the influence that a previous progression stage has on the future progression trajectory is encoded in its corresponding attention weight. 

\subsection{The Phased Attention Mechanism}
\label{Sec33}
% Equation 
To implement the attentive state dynamics, the attention weights in (\ref{eq2}) must be repeatedly updated (after each hospital visit) using variable-length sequences of data. Hence, we model the attention function \mbox{\footnotesize $\boldsymbol{\varphi}$} as an RNN that maps a patient's history to attention weights as follows:   
\begin{align}
\boldsymbol{h}_{m-1},.\,.\,.\,,\boldsymbol{h}_{1} &= \mbox{\textbf{\texttt{pLSTM}}}\,(\{(\boldsymbol{\tilde{X}}_{m-j},\Delta_{j,m})\}_{j=1}^{m-1}) ; \boldsymbol{\Theta}), \nonumber \\
\boldsymbol{e}_{j} &=\boldsymbol{w}^{T}\boldsymbol{h}_{j} + \boldsymbol{b},\, \forall j \in \{1,.\,.\,.\,,m-1\}, \nonumber \\
(\boldsymbol{\alpha}^m_1,.\,.\,.\,,\boldsymbol{\alpha}^m_{m-1}) &= \mbox{Softmax}(\boldsymbol{e}_{1},.\,.\,.\,,\boldsymbol{e}_{m-1}),
\label{eq3}
\end{align} 
where \mbox{\footnotesize $\Delta_{j,m} = t_{m-j+1}-t_{m-j}$}, \textbf{\texttt{pLSTM}} is a {\it phased} LSTM network \cite{neil2016phased}, \mbox{\footnotesize $\boldsymbol{w}$} and \mbox{\footnotesize $\boldsymbol{b}$} are the output layer parameters, \mbox{\footnotesize $\boldsymbol{h}_j$} is the hidden layer, and \mbox{\footnotesize $\boldsymbol{\tilde{X}}_{j} = [\boldsymbol{Y},\boldsymbol{X}_{j},t_j]$} is the input at time step \mbox{\footnotesize $j$}. Unlike traditional RNNs, a phased LSTM takes a timestamped sequence as an input, and performs updates at arbitrary points of time. Phased LSTMs can also handle asynchronously-sampled sequences, which is particularly important for EHR data as not all of the components of \mbox{\footnotesize $\boldsymbol{X}_m$} are necessarily measured in each hospital visits. Through phased LSTMs, we can update the attention weights with whatever follow-up data available at arbitrary time instances without the need for explicitly imputing missing observations. We call the attention mechanism in (\ref{eq3}) the {\it phased attention} mechanism. The PASS model is an attentive state space model that uses the phased attention mechanism. 

\begin{figure*}[t]
  \centering
  \includegraphics[width=5.5in]{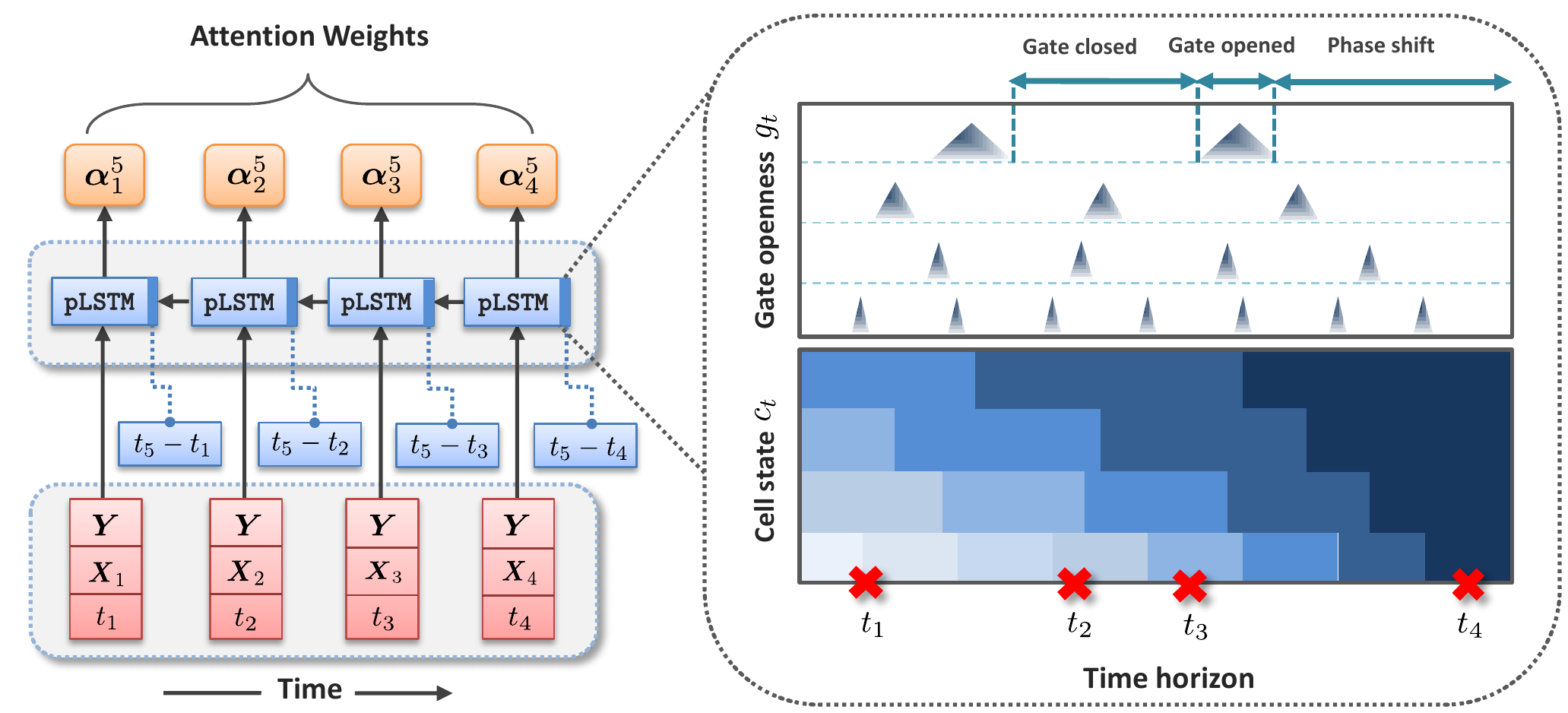}
  \captionof{figure}{\footnotesize Illustration of phased attention mechanism. {\bf Left:} Architecture of the phased attention network. The follow-up data (augmented with the static features) are fed in a reversed order into the phased LSTM, with reversed and shifted timestamps for the hospital visits. {\bf Right:} Illustration for the operation of the phased LSTM. The time gate \mbox{\footnotesize $\boldsymbol{g}_t$} of 4 neurons are depicted; each has different oscillatory parameters. The contents of the cell state \mbox{\footnotesize $\boldsymbol{c}_t$} decay as we go backwards in time, implying lesser attention for older hospital visits.}
	\label{Fig3}
\end{figure*}

In the \mbox{\footnotesize $m^{th}$} time step, phased attention operates by feeding the phased LSTM with the sequence \mbox{\footnotesize $\{\boldsymbol{\tilde{X}}_{j}\}_{j=1}^{m-1}$} in reversed order, with timestamps reversed and shifted by \mbox{\footnotesize $t_m$} as shown in Figure \ref{Fig3} (left). This allows all attention weights allocated to all previous state realizations (or equivalently, hospital visits) to be dynamically updated at every time step while preserving the relative time spacing between hospital visits. The phased attention mechanism in (\ref{eq3}) can be thought of as a continuous-time analogue of the reverse-time attention mechanism in \cite{choi2016retain}. 

The main difference between the phased LSTM model and the conventional LSTM model is the addition of a {\it time gate}, \mbox{\footnotesize $\boldsymbol{g}_t$}, which controls the updates to the LSTM cell state \mbox{\footnotesize $\boldsymbol{c}_t$} (and consequently the hidden layer \mbox{\footnotesize $\boldsymbol{h}_t$}). The opening and closing of the time gate \mbox{\footnotesize $\boldsymbol{g}_t$} for every neuron is controlled through an independent (continuous-time) rhythmic oscillation specified by 3 parameters (an oscillation period, a phase shift, and the ratio of the duration of the "open" phase to the full period) that can be learned from the data \cite{neil2016phased}. With every neuron having its own oscillatory parameters, the phased LSTM generates a continuum of possible updates that can be probed at arbitrary time instances as illustrated in Figure 3 (right). The updated equations of the phased LSTM can be found in \cite{neil2016phased}.   

\section{Learning and Inference}
\label{Sec4}
In this Section, we present the parameter learning and state inference algorithms for the PASS model. Throughout this Section, we continue with a single patient EHR record for the ease of notation.

{\bf Parameter learning}\,\,\, Let \mbox{\footnotesize $\boldsymbol{\Gamma}$} be the set of all PASS model parameters, i.e. \mbox{\footnotesize $\boldsymbol{\Gamma} = \{\boldsymbol{\Lambda},(\boldsymbol{\mu},\boldsymbol{\Sigma}),\boldsymbol{\Theta}\}$}, where \mbox{\footnotesize $\boldsymbol{\mu} = \{\boldsymbol{\mu}_z\}_{z=1}^D$} and \mbox{\footnotesize $\boldsymbol{\Sigma} = \{\boldsymbol{\Sigma}_z\}_{z=1}^D$} are the emission parameters. The {\it complete data} log-likelihood of an EHR record \mbox{\footnotesize $\boldsymbol{\mathcal{D}}$} and a state sequence realization \mbox{\footnotesize $\{\boldsymbol{Z}_m = z_m\}_m$} is given by:  
\begin{align}
\log(P(\boldsymbol{\mathcal{D}},\{z_m\}_{m=1}^M\,|\,\boldsymbol{\Gamma})) = \nonumber\\ \sum_{m=1}^M \log\left(\sum_{k=1}^{m-1}\boldsymbol{\alpha}_k^m\,\cdot\,\exp(\boldsymbol{\Lambda}\,(t_m-t_k))_{z_k,z_m}\right).
\label{eq2XWY}
\end{align}
Because the state sequence \mbox{\footnotesize $\{\boldsymbol{Z}_m\}_m$} is hidden, the complete data likelihood in (\ref{eq2XWY}) is inaccessible, and hence we resort to the Expectation-Maximization (EM) algorithm. The EM algorithm operates iteratively to update its guess of \mbox{\footnotesize $\boldsymbol{\Gamma}$}, where the $i^{th}$ iteration implements 2 steps: 
\begin{align}
\mbox{\bf E-step:}&\,\,\, \boldsymbol{Q}(\boldsymbol{\Gamma}\,|\,\boldsymbol{\hat{\Gamma}}^{(i)}) = \mathbb{E}\left[\,\log\left(P(\boldsymbol{\mathcal{D}},\{\boldsymbol{Z}_m\}_m)\right)\,\right], \nonumber \\
\mbox{\bf M-step:}&\,\,\, \boldsymbol{\hat{\Gamma}}^{(i+1)} = \mbox{argmax}_{\boldsymbol{\Gamma}} \boldsymbol{Q}(\boldsymbol{\Gamma}\,|\,\boldsymbol{\hat{\Gamma}}^{(i)}).
\label{eq3XWY} 
\end{align}
\begin{algorithm*}[t]
\caption{EM algorithm for learning the PASS model parameters}
\begin{algorithmic} 
\STATE {\bf Input:} EHR data $\boldsymbol{\mathcal{D}}$, initial guess $\boldsymbol{\hat{\Gamma}}^{(0)}$, state space $\{1,.\,.\,.,D\}$ and number of iterations $R$.
\STATE {\bf Output:} Parameter estimate $\boldsymbol{\hat{\Gamma}} = \{\boldsymbol{\hat{\Lambda}},(\boldsymbol{\hat{\mu}},\boldsymbol{\hat{\Sigma}}),\boldsymbol{\hat{\Theta}}\}$ 
\WHILE{$i \leq R$}
\STATE {\bf 1} \,$\bullet$\, $\{\hat{z}_m\}_m \leftarrow \mbox{{\bf Forward-Backward}}(\boldsymbol{\mathcal{D}}\,|\,\boldsymbol{\hat{\Gamma}}^{(i)})$
\STATE {\bf 2} \,$\bullet$\, $(\boldsymbol{\hat{\alpha}}^m_{1},.\,.\,.\,,\boldsymbol{\hat{\alpha}}^m_{m-1}) \leftarrow \boldsymbol{\varphi}((\boldsymbol{\tilde{X}}_{m-1},t_{m}-t_{m-1}),.\,.\,.\,,(\boldsymbol{\tilde{X}}_{1},t_{m}-t_1) ; \boldsymbol{\hat{\Theta}}^{(i)}),\, m \in \{1,.\,.\,.,M\}$
\STATE {\bf 3} \,$\bullet$\, $\boldsymbol{\hat{p}}_{m} \leftarrow \sum_{k=1}^{m-1}\boldsymbol{\hat{\alpha}}^m_k (e^{\Delta_k \, \boldsymbol{\hat{\Lambda}}^{(i)}})_{\hat{z},\hat{z}_k}\, m \in \{1,.\,.\,.,M\}$
\STATE {\bf 4} \,$\bullet$\, $\boldsymbol{\hat{\Theta}}^{(i+1)} \leftarrow \mbox{argmin} \left(-\sum_m \hat{z}_m\,\cdot\, \log(\hat{p}_m)\right)$
\STATE {\bf 5} \,$\bullet$\, $\boldsymbol{\hat{\Lambda}}^{(i+1)} \leftarrow \mbox{argmax} \log(P(\boldsymbol{\mathcal{D}}\,|\,\{\hat{z}_m\}_{m=1}^M,\boldsymbol{\Lambda})$
\STATE {\bf 6} \,$\bullet$\, $(\boldsymbol{\hat{\mu}}^{(i+1)},\boldsymbol{\hat{\Sigma}}^{(i+1)}) \leftarrow \mbox{argmax} \log(P(\{\boldsymbol{X}_m\}_m\,|\,\{\hat{z}_m\}_{m=1}^M)$
\STATE {\bf 7} \,$\bullet$\, $i \leftarrow i + 1$
\ENDWHILE
\label{alg1}
\end{algorithmic}
\end{algorithm*}
Algorithm 1 lists the steps involved in implementing the EM algorithm in (\ref{eq3XWY}). In Step 1, we first infer the hidden states via a message passing algorithm (described later) using the current guess of \mbox{\footnotesize $\boldsymbol{\Gamma}$}. Next, in Steps 2 and 3, we compute the attention weights associated with all hospital visit times, and then compute the transition probabilities using the formula in (\ref{eqy1}). In Step 4, the phased LSTM parameters are updated by optimizing the cross-entropy loss of the estimated transition probabilities and the inferred states. Maximum-likelihood is used to update the emission and transition parameters using the complete data likelihood obtained by plugging the inferred states into the expression in (\ref{eq2XWY}). To updated the Markov generator matrix \mbox{\footnotesize $\boldsymbol{\Lambda}$}, we use the {\it Expm} method in \cite{liu2015efficient}.  

\begin{figure*}[t]
  \centering
  \includegraphics[width=4.5in]{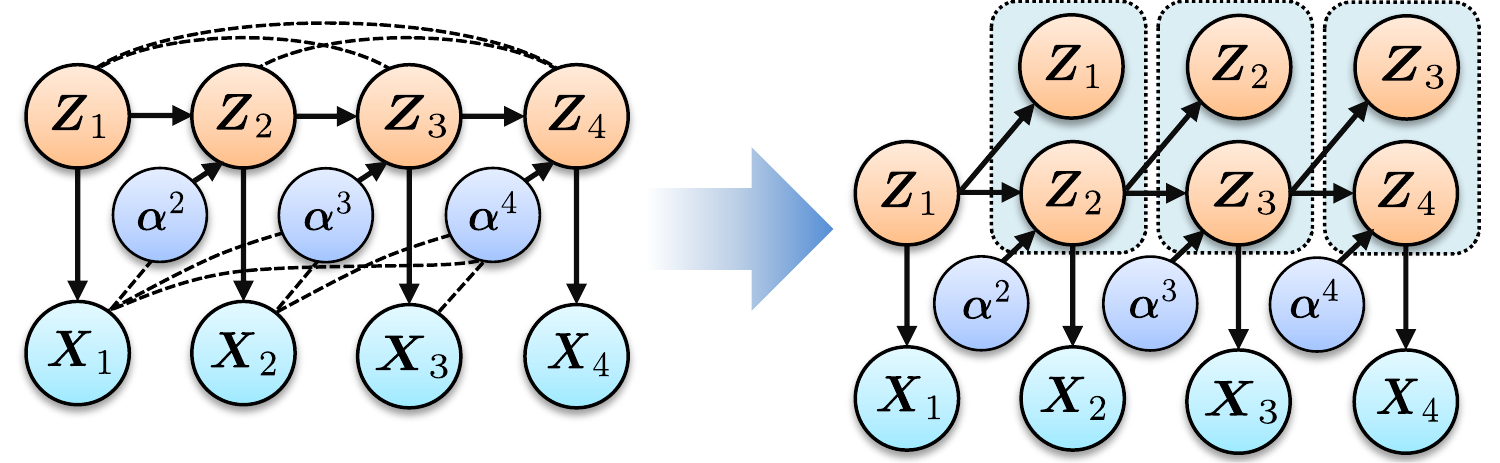}
  \captionof{figure}{\footnotesize Rearranged super states for the attentive model in Figure \ref{fig1c} with truncated attention and $Q=1$. The resulting graphical model (right) corresponds to a standard Hidden Markov model.}
	\label{Fig4}
\end{figure*}

{\bf State inference}\,\,\, One key advantage of the attentive state construction in (\ref{eq2XWY}) is that the attention weights explicitly quantify the importance of each past state to any given future state. Thus, efficient inference can be conducted by limiting the Markov blanket for every state variable to "important" past states with attention weights exceeding a certain threshold. Since attention weights already decline as we go back in time, we approximate the Markov blanket for every state $\boldsymbol{Z}_m$ by only considering the $Q$ most recent states $(\boldsymbol{Z}_{m-1},.\,.\,.,\boldsymbol{Z}_{m-Q})$. The resulting graphical model can be rearrange by lumping together every $Q$ consecutive states into one "super state" (as shown in Figure \ref{Fig4}), we retrieve a first-order Markovian dynamic Bayesian network (or equivalently, a higher-order Markov model \cite{murphy2002dynamic}), for which standard forward and backward message passing apply.   

%Finally, we note that in (\ref{eqy1}), we specified a finite-dimensional distribution over the embedded discrete process \mbox{\footnotesize $\{(\boldsymbol{Z}_m,t_m)\}_m$} rather than directly modeling the continuous-time process \mbox{\footnotesize $\boldsymbol{Z}(t)$}. This is because non-Markovian modeling of \mbox{\footnotesize $\boldsymbol{Z}(t)$} would require solving an intractable system of integral equations in order to compute the probability distribution of the discrete sequence \mbox{\footnotesize $\{\boldsymbol{Z}_m\}_m$} \cite{alaa2016hidden}. It is easy to show, using the {\it Kolmogorov extension Theorem}, that the model in (\ref{eqy1}) induces a consistent probability measure on \mbox{\footnotesize $\boldsymbol{Z}(t)$} \cite{kolmogorov2018foundations}. 

\section{Related Works}
\label{Sec2}

Previous works related to PASS fall into three areas: state space models of disease progression, RNN-based predictive models for healthcare applications, and (general-purpose) deep probabilistic models. In what follows, we discuss previous works in these three areas, and then conclude the Section by demonstrating the generality of the PASS model.

{\bf State space models of disease progression}\,\, Almost all existing models of disease progression are based on variants of the HMM model \cite{wang2014unsupervised,liu2015efficient,alaa2017learning}. Disease dynamics in such models are very easily interpretable as they can be perfectly summarized through a single matrix of probabilities that describes the transition rates among the different disease states. Markovian dynamics also greatly simplify inference because the model likelihood factorizes in a way that makes efficient forward and backward message passing possible \cite{murphy2002dynamic}. However, memoryless Markov models assume that a patient's current state $d$-separates her future trajectory from her clinical history. This renders HMM-based models incapable of properly explaining the heterogeneity in the patients' progression trajectories, which often results from their varying clinical histories or the chronologies (timing and order) of their experienced clinical events \cite{valderas2009defining}. This limitation is particularly crucial in complex chronic diseases that are accompanied with multiple morbidities. As discussed earlier, PASS addresses this limitation by creating memoryful state transitions that depend on the patient's entire clinical history. (In the Appendix material, we provide a detailed discussion on how PASS can better explain patient heterogeneity compared to Markov models.)

{\bf RNN-based predictive modeling for healthcare}\,\, Various RNN-based predictive models have been recently developed for healthcare settings; examples of such models include \textbf{\texttt{\small Doctor AI}} \cite{choi2016doctor}, \textbf{\texttt{\small L2D}} \cite{lipton2015learning}, and \textbf{\texttt{\small Disease-Atlas}} \cite{lim2018disease}. All those methods do not attempt to model a disease progression trajectory, but rather predict target clinical events on the basis of (discrete) sequential observations. Because of their black-box nature, none of these models can help understand the mechanisms underlying disease progression. 

There have been various attempts to create interpretable RNN-based predictive models using attention. The models in  \cite{choi2016retain} and \cite{ma2017dipole} use the {\it reverse-time attention} mechanism to learn visit-level and variable-level attention weights that explain the prediction of a target label through measures of variable importance. The phased attention mechanism proposed in Section \ref{Sec33} is a generalization of the reverse-time attention mechanism in \cite{choi2016retain} that can operate in continuous-time, and update the attention weights at irregularly-spaced and potentially incomplete observations. The main difference between the way attention is used in PASS and the way it is used in models like \textbf{\texttt{\small RETAIN}} \cite{choi2016retain} can be summarized as follows. PASS applies attention to the latent {\it state space}, whereas \textbf{\texttt{\small RETAIN}} applies attention to the observable {\it sample space}. Hence, the attention mechanism gives different types of explanations in the two models. In PASS, the phased attention mechanism interprets the hidden disease dynamics, and hence it provides an explanation for the mechanisms underlying disease progression. On the contrary, \textbf{\texttt{\small RETAIN}} uses attention to measure feature importance, and hence it only explains predictions, but does not explain the disease progression mechanisms.  

{\bf Deep probabilistic models}\,\, Most existing works on deep probabilistic models have focused on developed structured inference algorithms for deep Markov models and their variants \cite{krishnan2017structured,dai2016recurrent,karl2016deep,johnson2016composing}. All such models use neural networks to model the transition and emission distributions, but are limited to Markovian dynamics. Other works develop stochastic versions of RNNs for the sake of generative modeling; examples include variational RNNs \cite{chung2015recurrent}, SRNN \cite{fraccaro2016sequential}, and STORN \cite{bayer2014learning}. All such models augment stochastic layers to an RNN in order to enrich its output distribution. However, the transition and emission distributions in all these models cannot be decoupled, and hence their latent state representations would not lead to clinically meaningful identification of disease states. To the best of our knowledge, PASS is the first deep probabilistic model that provides both a clinically interpretable latent representation, and interpretable non-Markovian state dynamics. 

{\bf Generality of the attentive state space representation}\,\, For particular choices of the attention function in (\ref{eq2}), the attentive state space representation in (\ref{eqy1}) reduces to various classical models of sequential data. For instance, if \mbox{\footnotesize $\boldsymbol{\varphi}$} always sets \mbox{\footnotesize $\boldsymbol{\alpha}^m_{m-1}$} to 1 and all other weights to 0, then we retrieve an HMM. If the attention weights are binarized, then we retrieve a variable-order HMM \cite{willems1995context, begleiter2004prediction}. Furthermore, if the attention weights are fixed, then we recover an auto-regressive model. This is a powerful feature of our model as it implies that by learning the attention function \mbox{\footnotesize $\boldsymbol{\varphi}$}, we are effectively testing the assumptions of various commonly-used time series models in a data-driven fashion.

\section{Experiments}
\label{Sec5}
To validate the PASS model, we conducted a set of experiments using retrospective data for a longitudinal cohort of cystic fibrosis (CF) patients. CF is a life-shortening chronic condition that causes severe lung dysfunction, and is the most common genetic disease in Caucasian populations \cite{szczesniak2017phenotypes}. All experimental details are listed hereunder. 

Recall that, as stated in Section \ref{SSec1}, the main purpose of the PASS model is to simultaneously fulfill {\bf Goal A} (predicting individualized disease trajectories) and {\bf Goal B} (understanding disease progression mechanisms). Thus, in Sections \ref{SSec51} and \ref{SSec52}, we evaluate our model with respect to both goals.

{\bf Data description.}\,\, The dataset involved in the experiments was extracted from the UK CF registry, a database maintained by the UK CF trust\footnote{\url{https://www.cysticfibrosis.org.uk/the-work-we-do/uk-cf-registry/}}. Data was gathered from hospitals all over the UK, with 99$\%$ of patients consenting to their data being submitted, and hence the cohort is representative of the UK CF population. The dataset comprises longitudinal follow-ups for 10,263 patients over the period spanning between 2008 and 2015, with a total of 60,218 hospital visits. Each patient is associated with 90 variables, including the intake of 36 possible treatments, diagnoses for 31 possible comorbidities and 16 possible infections, FEV1 biomarkers, gender, and CF genetic mutations. 

\subsection{{\bf Goal A}: Predicting individual-level CF progression trajectories}
\label{SSec51}
{\bf Baselines.} We compared the predictive accuracy of PASS to the following models:
\begin{itemize}
\item {\bf MLP:} A multi-layer perceptron (MLP) classifier that is trained to sequentially predict the clinical events in 1 hospital visit given the observations in the prior visits. The MLP is trained on a static dataset that is created by unrolling the longitudinal follow-up data for all patients and treating every hospital visit as a separate data point.
\item {\bf HMM:} A standard continuous-time HMM model \cite{wang2014unsupervised,liu2015efficient} trained with the Baum-Welch EM algorithm. The number of HMM states was set via the Akaike information criterion (AIC). Similar to the PASS model, the observations \mbox{\footnotesize $(\boldsymbol{X}_{1},.\,.\,.,\boldsymbol{X}_{t})$} are modeled as Gaussian emission variables with state-specific mean and variance parameters.
\item {\bf RNN:} A standard LSTM network with 2 hidden layers of size 200. The follow-up data \mbox{\footnotesize $(\boldsymbol{X}_{1},.\,.\,.,\boldsymbol{X}_{t})$} was used as an input, and the output was defined as a set of (binary) labels designating the prediction targets at every hospital visit. A sigmoid transformation was applied to the top hidden layer. 
\item {\bf {\textbf{\texttt{RETAIN}}}:} An RNN-based reverse-time attention model proposed in \cite{choi2016retain}. To ensure a fair comparison with PASS and the standard RNN benchmark, we implemented the attention layer of {\small \textbf{\texttt{RETAIN}}} via an LSTM with 2 hidden layers of size 200, and restricted its architecture to generate only visit-level attention. (This is equivalent to the {\bf RNN-$\alpha_M$} benchmark in \cite{choi2016retain}.)
\end{itemize}

The baseline algorithms above are selected so as to highlight the added value of every modeling component in PASS. That is, an MLP only uses current information to predict the future clinical outcomes, whereas an HMM only looks 1 step back but provides a fully-fledged probabilistic model for disease progression. On the other hand an RNN capture more flexible dynamics (and is more memoryful) than an HMM but lacks interpretability, whereas {\small {\textbf{\texttt{RETAIN}}}} can provide explanations for its predictions, but does not explain the actual mechanisms of disease progression. 

{\bf Implementation of PASS.}\,\, We implemented the phased LSTM in Section \ref{Sec4} with 2 hidden layers of size 200 in order to match the model complexity of {\small {\textbf{\texttt{RETAIN}}}} and the standard RNN baseline. Hospital visits after which a death event happens within 3 years were explicitly labeled as the absorbing state \mbox{\footnotesize $D$} in our model. The number of states \mbox{\footnotesize $D$} was tuned via cross-validation to optimize the accuracy of predicting mortality events. PASS was implemented in {\small {\textbf{\texttt{Tensorflow}}}} \cite{abadi2016tensorflow}, and the phased attention layer was implemented via {\small {\textbf{\texttt{tf.contrib.rnn.PhasedLSTMCell}}}}. 

{\bf Prediction tasks and evaluation metric.}\,\, All models were used to sequentially predict the 1-year risk for 6 prognostic tasks of predicting 3 comorbidities and 3 lung infections that are common in the CF population. The comorbidities are Allergic bronchopulmonary aspergillosis (ABPA), diabetes and intestinal obstruction. The lung infections are Klebsiella Pneumoniae, E. coli and Aspergillus. We used the area under the ROC curve (AUC-ROC) with 5-fold cross-validation for performance evaluation, where error counts are taken over all patients and all hospital visits.

\begin{table*}[t]
\centering
\begin{tabular}{l||cccccc}           
 & {\tiny \bf ABPA} & {\tiny \bf Diabetes} & {\tiny {\bf I. Obstruction}} & {\tiny \bf K. Pneumoniae} & {\tiny \bf E. Coli} & {\tiny {\bf Aspergillus}}  \\
{\tiny \,\,\,\,\,\,\,\,\,\,\,\,\,\,\,\,\,\,\,\textbf{Model}} & {\tiny AUC-ROC} & {\tiny AUC-ROC} & {\tiny AUC-ROC} & {\tiny AUC-ROC} & {\tiny AUC-ROC} & {\tiny AUC-ROC}\\
\hline \hline
{\tiny \,\,\,\,\,\,\,\,\textbf{PASS}} & {\bf \mbox{\tiny 0.687 $\pm$ 0.022}} & {\bf \mbox{\tiny 0.771 $\pm$ 0.012}} & {\bf \mbox{\tiny 0.577 $\pm$ 0.018}} & {\bf \mbox{\tiny 0.718 $\pm$ 0.026}} & {\bf \mbox{\tiny 0.701 $\pm$ 0.019}} & {\bf \mbox{\tiny 0.640 $\pm$ 0.011}} \\ 
{\tiny \,\,\,\,\,\,\,\,\textbf{\texttt{RETAIN}}} & \mbox{\tiny 0.685 $\pm$ 0.026} & \mbox{\tiny 0.764 $\pm$ 0.014} & \mbox{\tiny 0.578 $\pm$ 0.014} & \mbox{\tiny 0.715 $\pm$ 0.031} & \mbox{\tiny 0.697 $\pm$ 0.015} & \mbox{\tiny 0.641 $\pm$ 0.010} \\ 
{\tiny \,\,\,\,\,\,\,\,\textbf{RNN}} & \mbox{\tiny 0.681 $\pm$ 0.016} & \mbox{\tiny 0.762 $\pm$ 0.021} & \mbox{\tiny 0.577 $\pm$ 0.010}  & \mbox{\tiny 0.719 $\pm$ 0.036} & \mbox{\tiny 0.696 $\pm$ 0.014} & \mbox{\tiny 0.641 $\pm$ 0.012}   \\  
{\tiny \,\,\,\,\,\,\,\,\textbf{HMM}} & \mbox{\tiny 0.666 $\pm$ 0.021} & \mbox{\tiny 0.755 $\pm$ 0.031} & \mbox{\tiny 0.551 $\pm$ 0.014} & \mbox{\tiny 0.689 $\pm$ 0.021} & \mbox{\tiny 0.665 $\pm$ 0.013} & \mbox{\tiny 0.620 $\pm$ 0.009}   \\   
{\tiny \,\,\,\,\,\,\,\,\textbf{MLP}} & \mbox{\tiny 0.657 $\pm$ 0.036} & \mbox{\tiny 0.751 $\pm$ 0.056} & \mbox{\tiny 0.553 $\pm$ 0.024} & \mbox{\tiny 0.685 $\pm$ 0.052} & \mbox{\tiny 0.656 $\pm$ 0.018} & \mbox{\tiny 0.601 $\pm$ 0.012} \\   
\end{tabular}
\caption{\footnotesize Performance of the different competing models for the 6 prognostic tasks under consideration.}
\label{Table2}
\end{table*}

{\bf Results.} The AUC-ROC performance of all models on the 6 prognostic tasks under consideration is provided in Table \ref{Table2}. We note that CF is a very complex disease, for which patients encounter various possible comorbidities and are prescribed a wide variety of possible treatments. This leads to each patient having a very rich clinical history that influence their outcomes. Because HMMs fail to properly integrate the patients' rich clinical histories into the state dynamics, they displayed modest predictive performance on the 6 prognostic tasks. As we can see in Table \ref{Table2}, the HMM model did not provide any significant improvement over the static MLP model in any of the prognostic tasks. On the contrary, RNN-based model provided significant improvements over the static MLP model on all of the 6 tasks. The results in Table \ref{Table2} show that the predictive accuracy of PASS is comparable to that of the RNN-based models. Note that the standard RNN model issue predictions without modeling disease progression, and hence it does not offer any interpretation benefit, whereas \textbf{\texttt{RETAIN}} provides explanations only in the form of measures of variable importance. PASS, however, explicitly models the CF physiology (in terms of its latent progression stages), and hence it ensures the interpretational and modeling benefits of an HMM while maintaining the predictive accuracy of an RNN-based predictive model.

\begin{figure*}[t]
  \centering
  \includegraphics[width=5.5in]{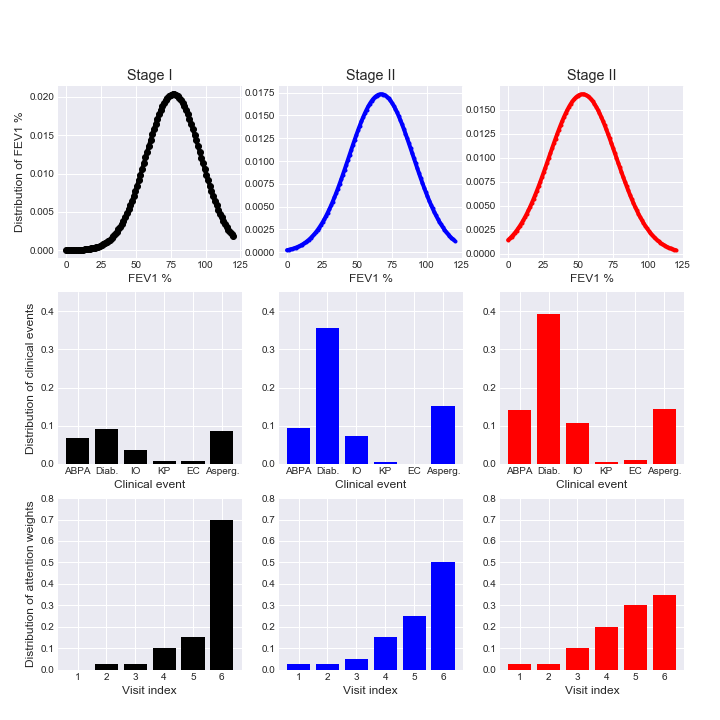}
  \captionof{figure}{\footnotesize Depiction for the distributions of clinical observations and generated attention in the 3 learned progression stages.}
	\label{Fig5}
\end{figure*}

\subsection{{\bf Goal B}: Understanding disease progression mechanisms}
\label{SSec52}

In this Section, we show that the PASS model successfully extracts meaningful clinical knowledge about the CF progression mechanisms. We also show how the PASS model parameters can inform clinical practitioners about the progression mechanisms for individual patients.  

{\bf CF progression stages.}\,\, Unlike many other chronic diseases, current clinical guidelines do not provide classifications for the progression stages of CF \cite{szczesniak2017phenotypes}. In a completely unsupervised fashion, the PASS model successfully learned \mbox{\footnotesize $D=3$} progression stages (Stages {\bf I}, {\bf II} and {\bf III}) that corresponded to clinically distinguishable levels of CF severity. The learned baseline Markov generator matrix for the transition rates among the 3 stages is given by:
\begin{align}
\boldsymbol{\hat{\Lambda}} = \begin{bmatrix}
\, -0.0578 & 0.0578 &  0  \, \\
\, 0  & -0.0691 & 0.0691 \, \\
\, 0 &  0 & 0 \,
  \end{bmatrix}. 
\label{eq00X2}
\end{align}  
From the baseline transition rates in (\ref{eq00X2}), it follows that the average occupancy of a patient in Stage {\bf I} is $\frac{1}{0.0578} = 17.30$ years, and the occupancy in Stage {\bf II} is around $\frac{1}{0.0691} = 14.47$ years. That is, a typical patient who is born with CF progresses to Stage {\bf II} by adulthood, before reaching the terminal stage by the age of 31. These figures match the survival rates in CF populations, where the median lifetime is known to be as low as 40 years of age \cite{mccarthy2013cf}. Note that the baseline generator matrix in (\ref{eq00X2}) only describes population-level rates of progression: individual variability among patients are captured via the patient-specific attention weights.  

The FEV1 $\%$ biomarker is the main spirometric measure of lung function that is currently used to guide clinical and therapeutic decisions. In order to check that the learned progression stages correspond to different levels of disease severity, we plot the estimated emission distribution for the FEV1 $\%$ biomarker in Stages {\bf I}, {\bf II} and {\bf III} in Figure \ref{Fig5}. As we can see from the emission distributions in Figure \ref{Fig5}, the mean values of the FEV1 biomarker in each stage were $87\%$, $65\%$ and $36\%$, respectively. This coincided with the current practice guidelines for referring critically-ill patients to a lung transplant, which recommends a transplant for patients with FEV1 $< 30\%$, monitoring for a transplant for patients with FEV1 ranging from $30\%$ to $80\%$, and no transplant for patients with FEV1 above $80\%$ \cite{braun2011cystic}. Thus, the learned progression stages can be translated into actionable information for clinical decision-making. We also plot the emission probabilities (parameter of the Bernoulli distribution) for the 3 comorbidities in Table \ref{Table2} at every stage. We can also see that the incidences of comorbidities increase significantly in the more severe Stages {\bf II} and {\bf III} as compared to Stage {\bf I}.

In Figure \ref{Fig5}, we also obtain the maximum a posterior inferences for the progression stages of every individual patient as described in Section \ref{Sec4}, and plot the average attention weights assigned to the patients' last 6 hospital visits in every progression stage. We found that the attentive dynamics tend to be less relevant for patients in Stage {\bf I}, where most of the attention is allocated to the most recent visit. Memory starts getting more important in Stages {\bf II} and {\bf III}, where the attention weights allocated to older hospital visits gets higher. This can be explained by the fact that patients in Stages {\bf II} and {\bf III} are more likely to have been diagnosed with more comorbidities in the past, and hence more segments of their clinical history matters for predicting their outcomes.  
 
{\bf Significance of attention weights for individual patients.}\,\, How can clinicians interpret and make use of the generated attention weights for the patient at hand? One important way to utilize the attention weights is to reason about the effect of different treatment decisions and how their outcomes are impacted by the patient's history. For the sake of illustrating this point, in Figure \ref{Fig6} we pick an out-of-sample patient who has repeatedly visited the hospital over the years 2012, 2013 and 2014. We see the attention weights generated by the PASS model can inform the clinician about the potential efficacy of the Ivacaftor treatment (a gene targeted therapy \cite{wainwright2015lumacaftor}) prescribed for this particular patient in the year 2014 by predicting the risk of progression to Stage {\bf III} of lung function severity by the year 2015.  

\begin{minipage}{0.425\textwidth}
The inferred progression stage for the patient was Stage {\bf II} for all of the 3 hospital visits. We toggle the patient's follow-up data vector \mbox{$\boldsymbol{X}_3$} (in year 2014) to let the variable indicating the prescription of the Ivacaftor drug be once set to 0 and once set to 1, and compute the probability of progressing to Stage {\bf III} by the year 2015 in each case. We found that assigning Ivacaftor treatment to this patients is actually associated with an elevated risk of progressing to the severe Stage {\bf III} within 1 year. By inspecting at the attention weights, we found that most attention is assigned to the most recent visit when Ivacaftor treatment is not prescribed, but the highest attention is paid to the follow-up data in 2012 when Ivacaftor is prescribed.       
\end{minipage}
\hfill
\begin{minipage}[h]{0.55\textwidth}
  \centering
  \includegraphics[width=3 in]{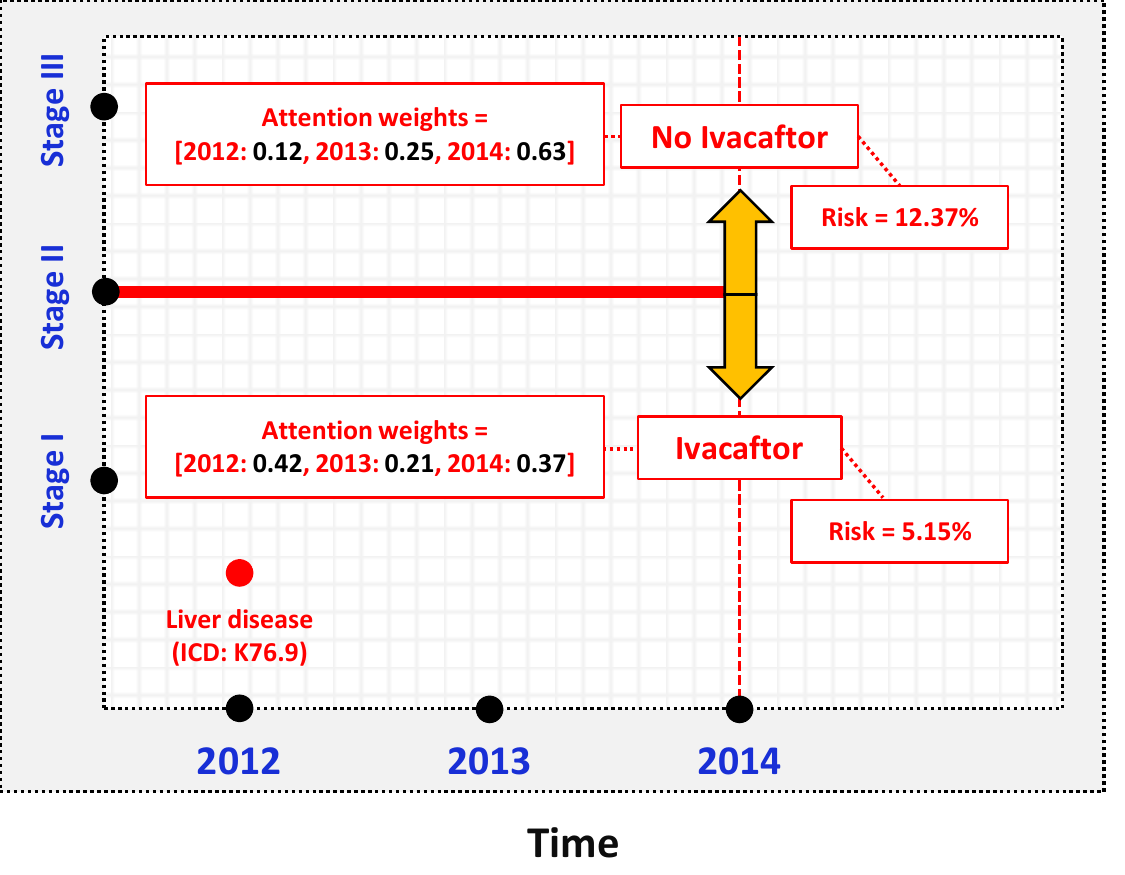}
  \captionof{figure}{\footnotesize Usage of the PASS attention weights to reason about treatment decisions.}
	\label{Fig6}
\end{minipage} 

The patient's follow-up data in year 2012 included a diagnosis with Liver disease. Hence, the elevated risk of progression upon taking the Ivacaftor treatment may be linked to its side effects concerning liver function complication, which may exacerbate the patient's liver disease. The PASS model altered the state dynamics to take into account the 2-year old follow-up data that is only important in determining the state dynamics conditional on prescription of an Ivacaftor treatment. 

\newpage
\section*{Appendix}
\subsection*{Detailed structure of the EHR data}
\begin{figure*}[h]
  \centering
  \includegraphics[width=5.5in]{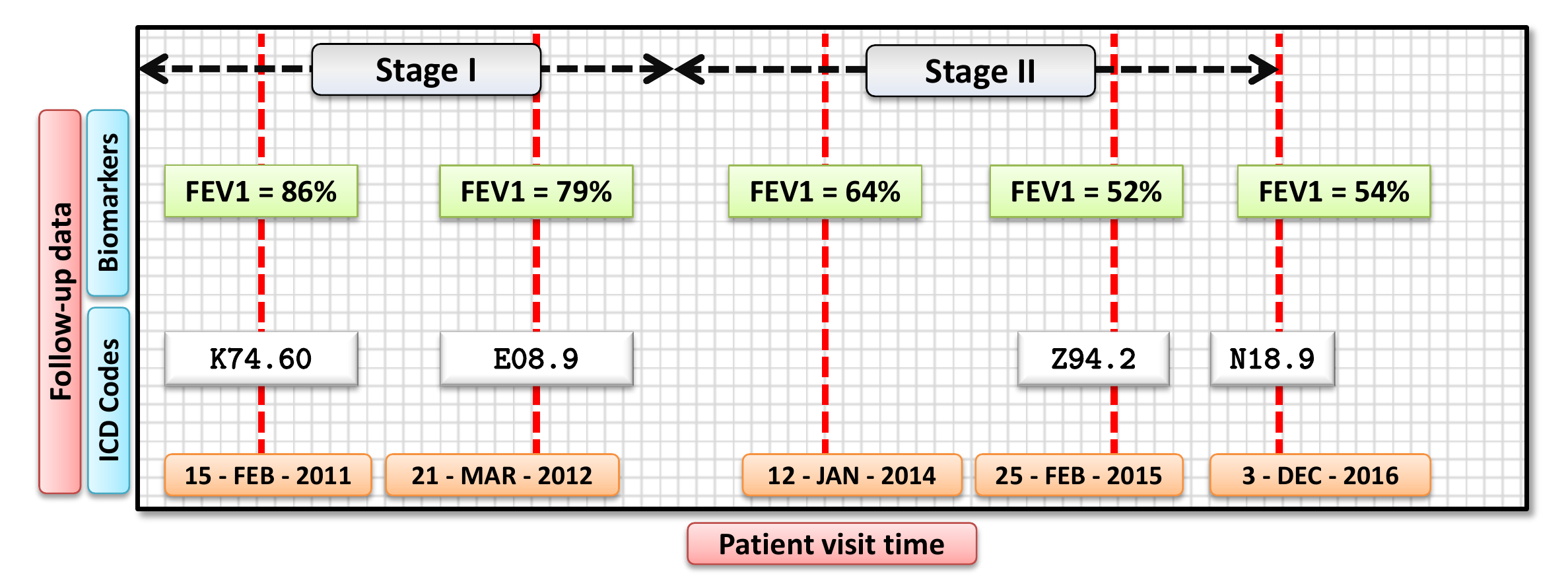}
  \caption{\footnotesize Illustration for the structure of the EHR data.} 
	\label{ApFig1}
\end{figure*}

A patient's EHR record, denoted as \mbox{\footnotesize $\boldsymbol{\mathcal{E}}$}, is a collection of timestamped follow-up data gathered during repeated, irregularly-spaced hospital visits, in addition to static features of the patient (e.g., genetic variables). We represent a given patient's EHR record as follows:  
\begin{equation}
\boldsymbol{\mathcal{E}} = {\annotate{$\{\boldsymbol{Y}\}$}{$\mbox{\footnotesize \bf Static features}$}} \cup \{(\boldsymbol{X}_{m},{\annotate{$t_{m}$}{$\mbox{\footnotesize \bf Visit times}$}})\}^{M}_{m=1},\,\, \mbox{\bf Follow-up data}\,\, \boldsymbol{\Longrightarrow}
\boldsymbol{X}_{m} = \Big[\,\, {\annotate{$\boldsymbol{u}_{m}$}{$\mbox{\bf \footnotesize Treatments}$}}\,\,\,,\,\,\,\,\,\,\,\,\,{\annotate{$\boldsymbol{C}_{m}$}{$\mbox{\,\,\,\bf \footnotesize Anchors}$}}\,\,\,,\,\,\,\,\,\, {\annotate{$\boldsymbol{O}_{m}$}{$\mbox{\footnotesize \,\,\,\,\,\,\,\,\,\,\,\,\, \bf Observations}$}} \,\,\Big], \nonumber
\end{equation}    
where \mbox{\footnotesize $\boldsymbol{Y}$} is the static features' vector, \mbox{\footnotesize $\boldsymbol{X}_{m}$} is the follow-up data collected in the \mbox{\footnotesize $m^{th}$} hospital visit, \mbox{\footnotesize $t_m$} is the time of the \mbox{\footnotesize $m^{th}$} visit, and \mbox{\footnotesize $M$} is the total number of hospital visits. (The time-horizon \mbox{\footnotesize $t$} is taken to be the patient's chronological age.) The follow-up data vector \mbox{\footnotesize $\boldsymbol{X}_{m}$} comprises three components:
\begin{itemize}
\item {\bf \underline{Treatments indicator} (\mbox{\footnotesize $\boldsymbol{u}_{m} \in \{0,1\}^U$}):} A binary vector indicating the prescription of (a subset of) \mbox{\footnotesize $U$} possible treatments to the patient during the \mbox{\footnotesize $m^{th}$} hospital visit. 
\item {\bf \underline{Anchor findin}g\underline{s} (\mbox{\footnotesize $\boldsymbol{C}_{m} \in \{0,1\}^K$}):} A binary vector indicating the presence of concrete diagnoses (i.e., ICD or HCPCS codes \cite{blumenthal2010meaningful}) for \mbox{\footnotesize $K$} distinct comorbidities that may co-occur with the target disease. 
\item {\bf \underline{Clinical observations} (\mbox{\footnotesize $\boldsymbol{O}_{m} \in \mathbb{R}^O$}):} A set of laboratory-measured biomarkers that reflect the severity of the target disease.
\end{itemize}
An EHR dataset \mbox{\footnotesize $\mathcal{D}$} is an assembly of records for \mbox{\footnotesize $N$} patients, i.e., \mbox{\footnotesize $\mathcal{D} = \{\mathcal{E}^{(i)}\}_{i=1}^N$}. Figure \ref{ApFig1} provides an illustration for the structure of the EHR data.

\newpage
\subsection*{Comparison between Markovian and attentive state dynamics}
Because Markovianity simplifies inference, most existing models of disease progression are based on HMMs (e.g., \cite{alaa2017learning}, \cite{liu2015efficient}, and \cite{wang2014unsupervised}). However, memoryless Markovian dynamics hinder a model's capacity for "explaining" individual-level progression trajectories. This is because under a Markov model, all patients at the same stage of progression would have the same expected future trajectory, irrespective of their potentially different individual clinical histories (i.e., the timing and order of treatments and comorbidities). That is, a Markov model captures the {\it population-level} transition rates among progression stages, but explains away {\it individual-level} variations in progression trajectories through the randomness of the transition probability \mbox{\footnotesize $P(\boldsymbol{Z}_m\,|\,\boldsymbol{Z}_{m-1},t_m-t_{m-1})$}. This can render Markov models highly misleading since clinical actions are taken on an individual basis. % clinical history only infers current stage, and the current stage explains everything
 
{\bf Interpreting the attentive state dynamics}\,\, Now we use an illustrative example to show how a clinician can interpret the attentive state dynamics for individual patients, and highlight the information that is missed by Markovian dynamics but can be captured via attentive dynamics.  

\begin{minipage}{0.515\textwidth}
In Figure \ref{Fig2OOO}, we display exemplary progression trajectories for 2 chronic kidney disease (CKD) patients through the unrolled graphical model of the DBN in (\ref{eq01}). With a slight abuse of graphical model notation, we let the thickness of the arrows connecting states be proportional to the attention weights generated for predicting the state transition in the third hospital visit. Patients A and B have identical trajectories at the first 2 visits (both are diagnosed with hypertension and are in the same progression stages), with one exception being that Patient A is administered a medication for hypertension (ACE inhibitors) in the first visit. Patient B transits from stage 2 to stage 3 CKD because of hypertensive renal complications, whereas patient A stays in stage 2 thanks to the medication. The attentive model can capture the difference between the 2 trajectories by paying attention to the first visit for Patient A (when the medication was prescribed), and little attention to the same visit for Patient B.   
\end{minipage}
\hfill
\begin{minipage}[h]{0.475\textwidth}
  \centering
  \includegraphics[width=2.5 in]{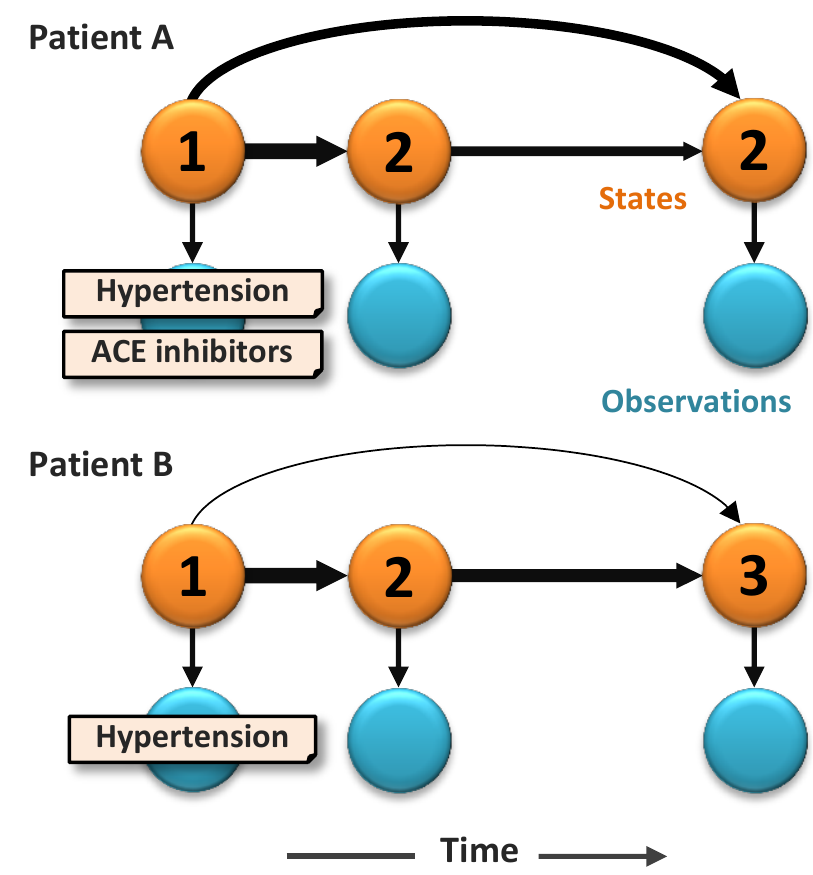}
  \captionof{figure}{\footnotesize Depiction for the attentive state dynamics.}
	\label{Fig2OOO}
\end{minipage} 

By visually inspecting the attention weights assigned to past states of each individual patient, clinicians can interpret the decreased risk for Patient A (compared to Patient B) to be a result of the clinical events that Patient A encountered in the first visit (i.e., administration of ACE inhibitors). On the contrary, a memoryless Markov model would not be able to distinguish the different trajectories that Patients A and B exhibit as both patients are in Stage 2 CKD during the second visit. 

\newpage
\nocite{*}
\bibliography{iclr2019_conference2}

\end{document}